# Generating *One*-Anaphoric Expressions: Where Does the Decision Lie?

Robert Dale
Microsoft Institute of Advanced Software Technology
North Ryde, Sydney
New South Wales 2113, Australia
rdale@microsoft.com

## Abstract

Most natural language generation systems embody mechanisms for choosing whether to subsequently refer to an already-introduced entity by means of a pronoun or a definite noun phrase. Relatively few systems, however, consider referring to entites by means of *one*-anaphoric expressions such as *the small green one*. This paper looks at what is involved in generating referring expressions of this type. Consideration of how to fit this capability into a standard algorithm for referring expression generation leads us to suggest a revision of some of the assumptions that underlie existing approaches. We demonstrate the usefulness of our approach to *one*-anaphora generation in the context of a simple database interface application, and make some observations about the impact of this approach on referring expression generation more generally.

## 1 Introduction

Anaphoric reference to an entity previously mentioned in a discourse can be carried out using any of a number of different strategies: in particular, pronominal anaphora, *one*-anaphora and definite noun phrase anaphora may each be used in appropriate discourse contexts, as demonstrated in examples (1)–(3) respectively.

(1) a. John has a red jumper.

b. He wears *it* on Sundays.

(2) a. John has a red jumper and a blue one.

b. He wears *the red one* on Sundays.

(3) a. John has a red jumper and a blue cardigan.

b. He wears *the jumper* on Sundays.

A considerable amount of work in natural language generation has focussed on the problem of generating definite noun phrase anaphora; see, for example, Appelt [1985], Dale [1989], Reiter [1990], Dale and Haddock [1991], Dale [1992], Reiter and Dale [1992] and Dale and Reiter [1995]. Work on the generation of pronominal anaphora is somewhat less developed, with researchers often falling back on some notion of focus as the prime determinant of whether pronominalisation is possible; the major problem here is coming up with an independently motivated notion of what it means to be 'in focus'. The generation of *one*-anaphoric expressions, however, has been virtually ignored, apart from some initial explorations in Davey [1979], Jameson and Wahlster [1982] and Dale [1992].

In this paper, we look at the generation of *one*-anaphoric expressions, and consider how such an ability might be situated within a system that can generate pronouns and definite noun phrases. This turns out to be problematic: in Section 3 we consider a number of relatively naive solutions, and abandon these in favour of an approach, argued for in Section 4, where the decision to refer to an entity by means of a *one*-anaphoric expression is largely determined at the level of discourse planning. Section 5 describes a simple database query system which generates responses involving *one*-anaphora making use of this insight, and Section 6 concludes by making some observations about the impact of this way of doing things on the task of referring expression generation in general.

# 2 A Model of Referring Expression Generation

In most conceptions of the problem of referring expression generation, the system is faced with the task of coming up with some linguistic expression that corresponds to some internal symbol, in such a way that the linguistic expression succeeds in identifying the intended referent for the hearer.

A very simple characterisation of the algorithm that underlies much of this kind of work is shown in Figure 1. The reality, even in implemented systems, is generally more complex, of course: pronouns may be used even if the intended referent is not in focus—see, for example, the centering algorithm of Grosz *et al* [1983]—and a definite noun phrase may be used even if the referent has not been mentioned before, or alternatively its form may be further constrained in some way by the structure of the discourse. However, these complications are not important for our present purposes.

A variety of approaches to the construction of definite noun phrases which are DISTINGUISHING DESCRIPTIONS are discussed in Dale and Reiter [1995]. The question this paper addresses is as follows: where, if at all, does the decision to use a *one*-anaphoric expression fit into this kind of algorithm?

# 3 Simple Approaches to Generating *One*-Anaphora

In this section we consider two relatively simple approaches to the generation of *one*-anaphoric expressions, and point out their limitations.

## 3.1 *One*-Anaphora as Syntactic Substitution

The phenomenon of *one*-anaphora is reasonably well discussed in the linguistics literature: in terms of X-bar theory, for example, the proform *one* is generally characterised as a substitute for an $\bar{\text{N}}$ constituent (see, for example, Radford [1981:94–95], McCawley [1988:185–186]); and the systemic literature provides considerable discussion of the nature of *one* as a substitute (see, for example, Halliday and Hasan [1976:89–98]). Although these treatments differ in a number of respects, both characterise effectively the same syntactic constraints on when *one*-anaphora is possible: the *one* form is seen to substitute for a head noun and some number of modifiers of that noun.

For the purposes of natural language generation, we could take this notion of substitution literally. Let's assume, for the moment, that a *one*-anaphor always has its antecedent in the previous clause. This is not always true, but the algorithm described here can be trivially extended to deal with other cases. The generation of *one*-anaphora can then be characterised as follows. Suppose $P$ is a set consisting of the noun phrase structures that appear in the previous clause:

- Given an intended referent $r$, determine the semantic content needed to identify this referent to the hearer.
- Work out the syntactic structure that realizes this semantic content; call this $s$.
- Compare $s$ against each $p \in P$, and look for common substructure starting at the head noun and working outwards; replace the largest common substructure found in $s$ by the form *one*.

So, given an antecedent noun phrase as in (4a) and a subsequent noun phrase as in (4b), we can substitute the *one* form to produce (4c), with the *one*-anaphor substituting for the $\bar{\text{N}}$ constituent *mouldy Germanic manuscript*.[1]

(4) a. [a [large [mouldy [Germanic [manuscript $_{\text{N}}]_{\bar{\text{N}}}]_{\bar{\text{N}}}]_{\bar{\text{N}}}]_{\text{NP}}$]

b. [a [small [mouldy [Germanic [manuscript $_{\text{N}}]_{\bar{\text{N}}}]_{\bar{\text{N}}}]_{\bar{\text{N}}}]_{\text{NP}}$]

c. [a [small [one $_{\bar{\text{N}}}]_{\bar{\text{N}}}]_{\text{NP}}$]

There are a number of problems with this approach—it allows too much and still too little; see [Dale 1992: 215–220] for some discussion—but most importantly, it seems a rather wasteful approach: since the commonality between the antecedent and the anaphor has something to do with shared *semantic* content, why go as far as working out the *syntactic* structure required

[1] We will fairly randomly switch between consideration of definite and indefinite *one*-anaphoric forms: for the purposes of the present discussion, any complications introduced by this aspect of discourse status appear to be orthogonal to the issues we are concerned with.

Given an intended referent $r$:
**begin**
  **if** $r$ is in focus **then** use a pronoun
  **elseif** $r$ has been mentioned in the discourse already
  **then** build a definite noun phrase
  **else** build an initial indefinite reference
**end**

Figure 1: A Skeletal Referring Expression Generation Algorithm

to realise the second NP? Syntactic substitution may be an appropriate way to characterise the behaviour of the *one* form in the abstract, but it is not a particularly sensible strategy for generating such forms.

## 3.2 *One*-Anaphora as Semantic Substitution

The above objection to the syntactic substitution approach already points to a better solution: look for shared structure at the semantic level. Suppose we have the semantic structure that corresponds to the noun phrase *a magenta Capri*, and suppose we have gone as far as to generate the semantic content that could be ultimately realised as the noun phrase *a reef-green Capri*. These semantic structures could be represented as shown in (5a) and (5b) respectively:[2]

(5) a.
$$\begin{bmatrix} \text{index: x1} & \\ \text{status:} & \begin{bmatrix}\text{given: } -\end{bmatrix} \\ \text{sem:} & \begin{bmatrix}\text{type: capri} \\ \text{colour: magenta}\end{bmatrix} \end{bmatrix}$$

b.
$$\begin{bmatrix} \text{index: x2} & \\ \text{status:} & \begin{bmatrix}\text{given: } -\end{bmatrix} \\ \text{sem:} & \begin{bmatrix}\text{type: capri} \\ \text{colour: reef_green}\end{bmatrix} \end{bmatrix}$$

We can compare these two structures, and allow *one*-anaphora to be used wherever the type attribute is the same, and zero or more of the other properties are shared. We then replace the shared elements by some null symbol, which the linguistic realiser takes as its cue to produce a *one*-anaphor:

(6)
$$\begin{bmatrix} \text{index: x2} & \\ \text{status:} & \begin{bmatrix}\text{given: } -\end{bmatrix} \\ \text{sem:} & \begin{bmatrix}\text{type: } \phi \\ \text{colour: reef_green}\end{bmatrix} \end{bmatrix}$$

Where more than just the type property is shared, it's convenient to use a more elaborate representation to make it easy to compare structures.[3] By identifying what it is that the antecedent NP and the anaphoric NP have in common at the level of semantics, we both avoid unnecessary work in building syntactic structure, and, as it happens, constrain more correctly the use of *one*-anaphoric forms: this method is elaborated further in [Dale 1992:220–226], and could probably be said to represent the state-of-the-art in approaches to the generation of *one*-anaphora. All the important elements of this 'semantic structure substitution' approach were, however, already present in the work of Webber [1979] and [Jameson and Wahlster 1982].

## 3.3 An Assessment of Semantic Structure Substitution

The approach suggested above provides us with a way of generating *one*-anaphoric expressions that fits into the general algorithmic structure we sketched in Figure 1; all that is required is that our algorithm maintain a distinction between determining the semantic content of a referring expression and the linguistic realisation of that semantic content, a separation that is

[2] These representations have their roots in the representations developed in Dale [1992]: see that work for further detail and justification.

[3] There are a number of related issues here concerned with just what properties have to be explicitly mentioned given the ordering constraints on adjectives. For example, if I say *a small one* after having mentioned *a large red book*, most hearers would assume that the second referent is red; but if I say *a blue one*, it is less clear whether or not the second referent is large. Domain-dependent knowledge seems to play a role here.

Given an intended referent $r$:
**begin**
  **if** $r$ is in focus **then** use a pronoun
  **elseif** $r$ has been mentioned in the discourse already
  **then begin**
    build the semantics for a definite noun phrase
    **if** there is shared structure with a previous noun phrase **then** elide it
  **end**
  **else begin**
    build the semantics for an initial indefinite reference
    **if** there is shared structure with a previous noun phrase **then** elide it
  **end**
**end**

Figure 2: A Revised Skeletal Referring Expression Generation Algorithm

useful for other purposes in any case.[4] We then complicate the second and third steps of the algorithm to check for the possibility of using a *one*-anaphoric construction once the semantic content has been determined. A revised version of the skeletal algorithm is shown in Figure 2.

However, there is something less than perfect about this approach. In our expository transition from syntactic substitution to semantic substitution, we have effectively shifted the decision to use *one*-anaphora further back in the generation process. It turns out that we can shift the decision further back still: since, as we argue below, *one*-anaphora is typically used to achieve a specific range of discourse functions, it makes sense to have the discourse planning stage of a generation system at least partly determine that *one*-anaphora should be used.

# 4 *One*-Anaphora as a Discourse Phenomenon

## 4.1 The Functions of *One*-Anaphora in Discourse

The basic idea proposed here is very simple. Observation suggests that *one*-anaphoric forms are used to achieve particular discourse functions; a common such function, for example, is to contrast two entities. It seems reasonable to suppose that, at the discourse planning stage, a generator will already know that it is contrasting two entities; but if the system knows that it is performing a contrast, then at that stage it should already be able to suggest that a *one*-anaphor may be used. In other words: why construct an elaborate mechanism to determine a semantic structure that can be subsequently elided if this is tantamount to rediscovering something the generator already knew?

The idea that *one*-anaphora is used in the context of particular discourse functions is not particularly new: Dahl [1985] and LuperFoy [1991:114–159] both discuss this aspect of *one*-anaphora at some length. LuperFoy's observations are closest to those that lie behind the view taken here. She suggests that uses of *one*-anaphoric forms correspond to three particular discourse functions: to contrast two sets of individuals, to denote a representative sample of a set introduced by the antecedent, and to refer to a new specimen of a type that is salient in the discourse. We focus here on the first two of these categories, exemplified in (7) and (8) respectively:

(7) a. John has a magenta Capri.
  b. Robert has *a reef-green one.*

(8) a. John has several cars.
  b. *The smallest one* is a Capri.

It may not have escaped the reader's notice that these two discourse functions are very close to what Rhetorical Structure Theory [Mann and Thompson 1987] would characterise as instances of the CONTRAST and ELABORATION relations.

[4] For example, it allows us to generate *a magenta Capri* and *a Capri which is magenta* as variants of the same basic semantic content.

## 4.2 How We Might Integrate *One*-Anaphora in Text Planning

How we might go about explaining the mini-discourses in examples (7) and (8) above from the point of view of text planning depends to some extent on whether we view them in the context of monologic discourse or dialog. The sentence pair in (7) is equally plausible both in cases where the two sentences are uttered by one speaker, and in cases where the second sentence is a response by a second speaker to the first speaker's utterance of the first sentence; the sentence pair in (8) is more likely to be uttered by one speaker.

Consider the monologic discourse cases first. It seems plausible to suggest that the sentences in each case are 'spoken as pairs'. In (7), the speaker utters the two sentences precisely in order to draw a contrast; in (8), the second sentence is only a coherent contribution to a discourse (even with the *one*-anaphor replaced by a full noun phrase) given the background provided by the first sentence.

In a natural language generation system which performs text planning, then, it is reasonable to view the contrast or elaboration that is being performed as the most important issue; the particular linguistic expressions constructed are subsidiary to these aims. Viewed in this way, it makes perfect sense for the text planner to PRESELECT some of the linguistic features of the utterances to be produced when the discourse relation has been decided upon.[5]

Clearly there are other forms of contrast than those realised by means of *one*-anaphors. The following two clauses, for example, exhibit a contrast:

(9) a. John hates Bill ...

b. ... but Fred likes him.

Even if we look at the specific case of contrasting two entities of the same type but with different properties, there are still linguistic devices other than *one*-anaphora that may be appropriate. Consider the alternatives in (10) and (11), for example:

(10) a. The African elephant has a long trunk.

b. The Indian elephant has a short *one*.

(11) a. The African elephant's trunk is long.

b. The Indian elephant's is short.

However, for present purposes we can focus on instances of contrast where it is like entities that are being contrasted, and assume for simplicity that *one*-anaphora is the only appropriate contrastive device available.

As suggested above, what this view does is to push the decision to use a *one*-anaphoric expression further back still in the generation process; but it could be objected that all we have done is to rename the problem, so that now, instead of asking when it is possible to use a *one*-anaphoric expression, we are left with the question of when it is possible or appropriate to draw a contrast. This is, nonetheless, progress, because it sites the decision in a far more appropriate place. Deciding when a contrast should be made is a much larger question that must be faced by any text planning system. Ultimately, the view taken here is that contrast is just one device that we use to produce coherent discourse: one way of characterising the general problem for a text planner is as the decision of what to say next, and here notions like topic maintenance and topic chaining are crucially important. Contrasting two clusters of information stored in a knowledge base is just another of these associative devices that can be used to build a coherent text on the basis of relations that reside in the underlying knowledge base.[6] The work of McKeown [1985], Hovy [1991] and Sibun [1991] are all different approaches to this more problem of coherence: although the mechanisms adopted in each case are different, they all require the underlying knowledge base to play a role in determining what can be said next in a coherent text.

# 5 Implementing *One*-Anaphora in Dialogue

The preceding discussion is still, of course, rather vague, and awaits a more fleshed-out the-ory of the interaction of discourse planning decisions with surface linguistic phenomena. In this section we try to make the idea a little more concrete by looking at the use of contrastive *one*-anaphoric forms in dialogue; we describe a simple implementation of a database query system, with discourse planning being carried out by a dialogue manager component whose responsibility is to produce cooperative responses.

[5] The use of the term 'preselect', a term from work in systemic approaches to generation, is deliberate. What we are arguing for here amounts to an interstratal preselection of lexicogrammatical features in the sense of Matthiessen and Bateman [1991:62–65].

[6] All of the immediate discussion has been in terms of the contrastive discourse function of *one*, but a similar story can be told for the set-elaborative function.

To provide a focus here, we will use one particular example to motivate the discussion:

(12) a. Is there a flight to Melbourne before 7am?

b. No, but there's *one at 715.*

In line with the approach we suggested above, the claim here is that the generation of the *one*-anaphoric form is best seen as a direct consequence of the fact that the dialogue manager produces a co-operative response by drawing attention to some entity whose properties contrast with those specified in the user's query. Below we show how this works.

## 5.1 An Overview of the System

We have a database of flight information that we want to interrogate, and we want to get back useful responses from that database. We assume the existence of a simple query interpreter that takes in sentences in natural language and produces from them database queries that can be applied against the database: the issues involved in mapping from natural language queries to queries that can be applied against a database are well-explored in the literature and not of direct relevance to our task here (see Androutsopoulos *et al* [1994] for a good overview of many of the issues).

The core element of the system is the Dialogue Manager, which takes the query and decides how to process it with respect to the underlying database. Dealing with the inclusion of appropriate values for unspecified attributes is handled by the Dialogue Manager (in the example above, note that the city from which the flight departs is left implicit in the query), as are a number of other inference-based mechanisms; see Bobrow *et al* [1977] for an early system that identified many of the issues here. In our implementation, we assume simple defaults to deal with these phenomena.

In cases where this initial query results directly in the return of appropriate information from the database, nothing much needs to be done; but in general, if co-operative responses are to be produced by such a system, the Dialogue Manager must reason about the speaker's plans and goals, and use this information in determining precisely what kinds of information are most useful, and in determining what content is required in order to respond co-operatively. Allen and Perrault [1980] and Kaplan [1983] discuss many of the issues that arise in this connection.

If applying the query straightforwardly results in a null response, the task of the Dialogue Manager is to relax the query in some appropriate manner and try again. As explained in Section 5.3 below, it is this relaxation process that gives rise to the identification of contrast. The Dialogue Manager then constructs a speech act specification as a response, and passes this to the linguistic realiser for output as a natural language response.

## 5.2 The Underlying Knowledge Base

The knowledge base used to answer queries in the manner shown in example (12) obviously has to contain, at a minimum, information about flight departure points and destinations, and the times of departure. A fully-fledged flight database would of course contain other information too, but these items are sufficient for our present purposes. The information relevant to the above query is represented in a Prolog knowledge base as shown in Figure 3. The basic ideas here are fairly straightforward: the domain is populated by entities, and these entities have properties. The type property specifies how the entity fits into some taxonomy; the other properties are either attributes with constants as values, as in starttime and name, or attributes with other entities as values, as in startpoint and endpoint.

## 5.3 The Dialogue Manager

The Dialogue Manager has the job of coordinating some response to the query. We can think of a query $Q$ as a set of query elements that correspond to the attributes of the objects we are interested in. The set of query

```
entity(qf400).
property(qf400, type, flight).
property(qf400, name, "QF400").
property(qf400, startpoint, s1).
property(qf400, endpoint, m1).
property(qf400, starttime, 0715).
property(qf400, endtime, 0830).

entity(s1).
property(s1, type, city).
property(s1, name, "Sydney").

entity(m1).
property(m1, type, city).
property(m1, name, "Melbourne").
```

Figure 3: The Prolog Knowledge Base

elements is defined by a QUERY FRAME; this is essentially the same mechanism that allows database query systems in the tradition of GUS [Bobrow *et al* 1977] to determine what information is required in order to complete a transaction. Given the content of a natural language query, the Dialogue Manager uses the information in this query, plus some other information inferred from the context, to populate the query frame. For each query element in the query frame, there is a variable and a set of associated constraints over that variable; for each set of constraints we maintain a record of whether those constraints are as specified in the initial query, or whether they have been relaxed; and we maintain a record of those values for variables which are specified in the query applied against the database (the GIVEN constraints) and those values which are obtained as a result of the query (the NEW constraints).

Once fully processed, the query in (12) above results in the query frame shown in Figure 4. From this query frame, we construct a Prolog query which can be applied against the underlying knowledge base. In the context of our present example, this results in the following Prolog query:

```
(13) ?- entity(E),
        property(E, type, flight),
        property(E, startpoint, s1),
        property(E, endpoint, m1),
        property(E, starttime, T1),
        T1 < 700,
        property(E, endtime, T2).
```

Given the database shown in Figure 3, this Prolog goal fails. This causes the Dialogue Manager to relax the constraints in the query frame. Ideally this relaxation is carried out by a sophisticated reasoning and plan recognition component: it would generally be unwise, for example, to relax the constraint on destination, although in some circumstances this might be exactly what is required. In the current model, we simply use a domain-dependent heuristic that first weakens the constraint on starttime. The resulting query frame is shown in Figure 5. This in turn results in a new Prolog query. As a result of applying this query against the database, we get values for anything not completely specified, resulting in the query frame shown in Figure 6. A direct result of this encoding of the information is that we can see which attributes of the entity found in the database were already specified in the initial query and which correspond to new information.

## 5.4 Generating a *One*-Anaphor

Given this information, planning a response which makes use of a *one*-anaphor is then trivial. To justify the use of *one*-anaphora, all we need to do is ensure that the 'principle term' (here, the fact that the found entity is a flight) is not relaxed; otherwise the head noun would have changed and we could not use *one*-anaphora. We then look down the elements of the query frame, determine for which elements the constraints had to be relaxed, and construct a semantics for the NP which is empty except for these relaxed constraints: it is precisely these relaxed constraints which correspond to the contrasted properties. The other elements can be ignored because they do not present new information.[7]

The semantic input for the realiser specifies that a negative response should be generated, and that a contrastive solution should be offered; the important part of this, which corresponds to the

[7]Note that, in the current example, we also retrieve information about the arrival time of the flight in question, although we do not make use of this here. This could be used to generate co-operative responses that provide additional information along the lines suggested by Kaplan [1983].

| Attribute | Variable | Status | Given | New |
|---|---|---|---|---|
| entity | E | initial | | |
| type | T | initial | T = flight | |
| starttpoint | C1 | initial | C1 = s1 | |
| endpoint | C2 | initial | C2 = m1 | |
| starttime | T1 | initial | T1 < 700 | |
| endtime | T2 | initial | | |

Figure 4: The Initial Query Frame

| Attribute | Value | Status | Given | New |
|---|---|---|---|---|
| entity | E | initial | | |
| type | T | initial | T = flight | |
| starttpoint | C1 | initial | C1 = s1 | |
| endpoint | C2 | initial | C2 = m1 | |
| starttime | T1 | relaxed | T1 < 0800 | |
| endtime | T2 | initial | | |

Figure 5: The Relaxed Query Frame

| Attribute | Value | Status | Given | New |
|---|---|---|---|---|
| entity | E | initial | | E = qf400 |
| type | T | initial | T = flight | |
| starttpoint | C1 | initial | C1 = s1 | |
| endpoint | C2 | initial | C2 = m1 | |
| starttime | T1 | relaxed | T1 < 0800 | T1 = 715 |
| endtime | T2 | initial | | T2 = 830 |

Figure 6: The Instantiated Query Frame

noun phrase description of the entity to be described, is as shown in the following attribute–value matrix:

(14)
$$\begin{bmatrix} \text{status:} & \begin{bmatrix}\text{given: } -\end{bmatrix} \\ \text{sems:} & \begin{bmatrix} \text{type: } \phi \\ \text{properties: } \begin{bmatrix}\text{starttime: } 0715\end{bmatrix}\end{bmatrix}\end{bmatrix}$$

This is then passed to the grammar; the empty type is taken as a cue by the grammar to produce a *one*-anaphoric form, resulting in:

(15) No, but there is *one at 715am.*

There are, of course, a range of other responses that could be generated, but the basic point should be clear: we have succeeded in generating an appropriate *one*-anaphoric response that provides a contrast with a previous specification, without any need to first construct an entire noun phrase semantics and then look for commonality that can be elided. The *one*-anaphor falls out of the query processing carried out by the Dialogue Manager.

# 6 Conclusions and Future Work

We have argued that *one*-anaphora is best viewed as a discourse phenomenon, and that a consequence of this is that the decision to use *one*-anaphora should, at least in part, be determined at the level of discourse planning. We have demonstrated how this can operate in the context of a simple database query system, where the Dialogue Manager's attempts to find a useful response from a database correspond to the relevant discourse planning operations.

At the outset, we asked how the generation of *one*-anaphora could be integrated into existing referring expression generation algorithms. These algorithms assume that they are given some symbol that corresponds to the intended referent, and then attempt to determine what content should be used to identify this intended referent. This model is incompatible with the approach taken to *one*-anaphora here, since the approach we have argued for lacks a distinct stage in the processing where the intended referent is only indicated by some internal symbol.

In order to integrate the generation of *one*-anaphora into these standard algorithms, the assumption that the referring expression generator is given nothing more to work with than the symbol that corresponds to the intended referent has to be abandoned, and the bandwidth of communication between the discourse planner and the referring expression generator increased: ideally, the referring expression generator is told not only what the intended referent is, but also what its function in the discourse is.

This is not such a radical idea. McDonald's [1980] work on referring expression generation within MUMBLE includes a facility whereby the expert system driving the generator can specify that a message element (i.e., an internal symbol corresponding to the intended referent) is 'ontologically of a sort that cannot be pronominalized' [1980:217]: this allows the expert system to specify that some information has to be expressed for descriptive, rather than purely referential, purposes. A similar broadening of the bandwidth is visible in McKeown's TEXT [McKeown 1985], where the text planning component can indicate to the linguistic realisation component that a particular entity is the focus of the utterance, resulting in pronominalisation; and the same idea finds expression in the use of the CENTRE attribute in Dale's EPICURE [1992:170–171]. The present work suggests that these devices can be seen as instances of a more general mechanism where the discourse purpose of a referring expression plays a role in how that referring expression is best realised.

Further work is required in order to determine how best to rearrange generation architectures to integrate these observations. By abandoning the traditional architectural division into strategy and tactics [Thompson 1977], systems based on systemic functional grammar (see, for example, [Matthiessen and Bateman 1991]) already allow sufficient flexibility to incorporate the mechanisms discussed here. However, the absence of distinct processing modules with well-defined interfaces between them is generally considered to make it more difficult to build practical systems which can be easily re-used and maintained. A question for further research is whether, taking the observations of this paper into account, we can characterise the required interactions between referring expression generation and other aspects of the generation task in such a way that modular systems can be built.

## Acknowledgements

Thanks are due to Mark Lauer and Adrian Tulloch for helpful comments on an earlier version of this paper, and to a large number of *Linguist* mailing-list readers who provided pointers to work on *one*-anaphora in the wider literature.